**Large language models and linguistic intentionality**

**Jumbly Grindrod, University of Reading**

**Published in *Synthese:* https://link.springer.com/article/10.1007/s11229-024-04723-8**

**Please cite published version.**

***Abstract***: *Do large language models like Chat-GPT or Claude meaningfully use the words they produce? Or are they merely clever prediction machines, simulating language use by producing statistically plausible text? There have already been some initial attempts to answer this question by showing that these models meet the criteria for entering meaningful states according to metasemantic theories of mental content. In this paper, I will argue for a different approach – that we should instead consider whether language models meet the criteria given by our best metasemantic theories of linguistic content. In that vein, I will illustrate how this can be done by applying two such theories to the case of language models: Gareth Evans' (1982) account of naming practices and Ruth Millikan's (1984, 2004, 2005) teleosemantics. In doing so, I will argue that it is a mistake to think that the failure of LLMs to meet plausible conditions for mental intentionality thereby renders their outputs meaningless, and that a distinguishing feature of linguistic intentionality – dependency on a pre-existing linguistic system – allows for the plausible result that LLM outputs are meaningful.*

**Acknowledgements**

Thanks go to the two anonymous referees, who provided insightful comments on the paper. I am especially grateful to Patrick Butlin, who took time to comment on and discuss the paper in detail. Thanks also go to Nat Hansen, J.D. Porter, and the audience of University of Reading Work in Progress presentation.

## 1. The question of meaningful usage

Reproducing the sounds or shapes of words is not sufficient for meaningful language use. To use some familiar examples from the philosophical literature, if an ant crawling on a beach happens to trace a line with the exact shape of "Nero played while Rome burned," it has not thereby meaningfully used the sentence (Putnam 1981). Similarly, if a parrot is trained to reproduce the sentence's phonology, it has not thereby meaningfully used the sentence (Bender and Koller 2020).

Should we group large language models (LLMs) with the ant and the parrot?[1] Should we view these models as using language in the same way that we do with one another, or should we view them as mere mimics of language use? Already, both within academia and popular discourse, we have seen two broad positions staked out in response to this question. The generous position is that LLMs have the requisite intelligence and whatever other necessary features to communicate in the fullest sense possible. This may be because they possess general intelligence (Bubeck *et al.* 2023), or because they are sentient, or because they are conscious, etc.[2] On the other hand, the skeptical position, voiced for instance by Bender and Koller (2020), is that LLMs are merely mimicking language use through next-word prediction. Just as a parrot does not meaningfully use a word when it mimics the noise a speaker would make in using it, LLMs are "stochastic parrots" that mimic the language use that it was trained on (Bender *et al.,* 2021).[3]

The wider discussion can often conflate a number of issues in a way that philosophers are trained to avoid. One way to precisify the issue is to focus specifically on the question of whether the outputs produced by LLMs are meaningful or not. In asking this question, we are setting aside questions of general intelligence, sentience, or consciousness. We are also setting aside much more closely-related questions regarding the possible speech acts performed by LLMs, or whether certain forms of pragmatic content (e.g. implicatures) can be generated by LLMs.

[1] Strictly, there is a distinction to be drawn between LLMs and chatbots that are powered by LLMs, not least because LLMs have many uses other than acting as a chatbot. To avoid verbosity, I will ignore this distinction for the purposes of this paper.

[2] Famously, Blake Lemoine declared that one of Google's earlier LLMs Lamda was sentient (Wertheimer, 2022), while Ilya Sutskever, one of the co-founders of OpenAI has previously tweeted "it may be that today's large neural networks are slightly conscious" (Cuthbertson, 2022).

[3] There are ways of finding a middle ground between these two positions, for instance by arguing that LLMs do possess certain core linguistic abilities but lack others, or are able to communicate certain kinds of content but not others. The Evansian picture I consider in section 4.1 is a middle-ground position of this kind insofar as it claims that while LLMs can use words meaningfully, there are certain linguistic practices – i.e. acting as a producer for a naming practice – that they cannot take part in.

Instead, we are asking whether the text generated by LLMs are instances of meaningful language use. Let's call this the *meaningful usage question*.[4]

How might we go about answering this question? One of the key topics in analytic philosophy of language and mind has been to produce a theory of what it takes for something to be meaningful. A theory of this kind is sometimes described as a *metasemantics*.[5] We can draw a broad distinction between two types of metasemantics: one that captures how a mental state has meaning (what I will call a *mental metasemantics*) and one that captures how a linguistic item has meaning (what I will call a *linguistic metasemantics*). So one approach we can take to answering the meaningful usage question is to consider whether LLMs meet the conditions prescribed by our best metasemantic theory. Doing so will naturally require that we pay careful attention to the inner workings of LLMs. However, we should also keep in mind a point by Cappelen and Dever (2021), that the question may not be settled by inspecting the inner workings of an artificial system. If externalism about meaning is true, then whether a given entity counts as meaningful partly depends on features external to that entity. So in considering the meaningful usage question, we may also need to consider the way that LLMs are related to their environment.

Still, externalism only entails that external features play some role in meaning determination; it does not mean that we can ignore internal features altogether. And in the case of LLMs, it is the way that LLMs have been constructed and the way they process linguistic data that leads many to the skeptical position. Given this, it will be important in answering the meaningful usage question that attention is paid to how LLMs function.

---

[4] Lederman and Mahowald (2024) distinguish between "basic meaning" and "derived meaning". We could then precisify this question by asking whether LLM outputs are meaningful in either sense. However, they *stipulate* that only tokens created by agents with beliefs, desires, and intentions possess basic meaning, By contrast, the motivation of this paper is to consider whether LLMs that lack such mental states could nevertheless meaningfully use words in a non-derivative sense.

[5] Similarly, Stalnaker (2017) uses "foundational semantics" for theories that capture the facts that give rise to meaningful linguistic tokens. He contrasts this project with descriptive semantics, which seeks to assign meanings to each linguistic expression. The same distinction can be drawn by distinguishing between metasemantics and semantics (analogous to the meta-ethics/ethics distinction).

The plan for this paper is as follows. In the next section, I will provide an overview of how LLMs work, from simpler static systems through to the transformer architecture that is at the heart of all state-of-the-art LLMs. In doing so, particular emphasis will be given to the theoretical background that motivates LLMs, namely *distributional semantics*. Doing so is important as it provides a basis to respond to certain forms of argument for skepticism about meaningful LLM usage. In section 3, I will then turn to what I take to be the dominant approach to the question of meaningful usage i.e. considering whether LLMs meet the conditions of intentionality according to mental metasemantic theories. I will focus particularly on Coelho Mollo and Millière's (2023) recent arguments that LLMs can be meaningful users. I will argue that it is not plausible that LLMs can be treated as sources of mental intentionality. In section 4, I will then make the case that LLMs are better understood through the lens of linguistic metasemantic theories. After all, such theories attempt to capture meaningful language usage, and this is precisely what we are considering in asking the meaningful usage question. I will make the case that LLMs are meaningful language users by applying two distinct metasemantic theories: Evans' (1982) account of naming practices and Millikan's (1984, 2004, 2005) teleosemantic theory of linguistic meaning.[6]

## 2. The construction and theoretical background of large language models

LLMs have their origins within an approach to meaning known as *distributional semantics* (Erk, 2012; Lenci, 2018; Boleda, 2020; Grindrod 2023). This approach has a history stretching back to the likes of John Firth (1957) and Zellig Harris (1954) and is underpinned by the so-called *distributional hypothesis:* that the meaning of a word can be represented by its distribution across a text or corpus.[7] While this may look like a highly controversial principle to base an approach to meaning on, the most plausible gloss of distributional semantics treats this as a *hypothesis* that is worth testing to see if it bears fruit, rather than an unquestioned principle or a conclusion to some theoretical argument.[8] And in that respect, the recent success that we see in large language models constitutes a partial vindication of the approach.

[6] For the purposes of this paper, "meaningful", "intentional" and "representational" will be treated as synonymous. There are sometimes good reasons to distinguish between these notions, such as when we are considering types of meaning that may not be representational – expressivist conceptions of meaning could be understood in this way. However, such conceptions are not the concern of this paper.
[7] See: (Lenci, 2008; Grindrod 2023) for a number of different formulations of the distributional hypothesis.

One common way of employing distributional semantics is to represent the distribution of a word using a vector in a high-dimensional space. Distributional semantics then predicts that the proximity between two words will correspond with their meaning similarity. But how could a vector represent a word's distribution? Initially, a "count" approach was common, whereby each component of a word vector would represent how often that word co-occurs with some other word (Baroni, Dinu and Kruszewski, 2014). For example, a word like "dog" might have a component in its vector that stands for how often "dog" co-occurs with "bark". Even in this brief description of the simplest form of count approach, a number of further questions arise: how close must two words be to count as co-occurring? How many components does each vector need? Do the various inflections of a word get treated as the same word? These are all treated as parameters of the approach that can be tested in terms of how their adjustments affect evaluation performance (Kiela and Clark, 2014). There are also various innovations to the basic count approach that have been explored extensively. For instance, a common operation has been to use dimensionality reduction to reduce the size of the model and to distil the information to its core features.

A great deal of progress was made with the count-based approach, notably with breakthrough models such as *Latent Semantic Analysis* (Landauer and Dumais, 1997) and *Hyperspace Analogue to*

---

[8] It has sometimes been suggested that the distributional hypothesis is a natural outcome of the use-based theory of meaning associated with the later Wittgenstein (Lenci, 2008, p. 7; Erk, 2012, p. 635). Recruiting Wittgenstein to the cause is questionable for a couple of reasons, however. First, Wittgenstein is sometimes interpreted as a quietist about meaning, and such an interpretation would find distributional semantics as objectionable as any other theory of meaning. Second, it is clear from (Wittgenstein, 1953, § 2 ff.) that Wittgenstein envisioned use as understood both in terms of linguistic and non-linguistic activities. From this perspective, usage as represented by a corpus will be impoverished.

*Language* (Lund and Burgess, 1996), which are still used in a number of experimental contexts today. But there was a leap in progress in the performance of distributional models when Mikolov *et al.* (2013) introduced their Word2vec algorithms as a way of producing vectors through the use of a small neural network. The network is trained to predict a missing set of words given some text as input. Effectively, this means taking a one-hot vector as input (i.e. a vector whose components correspond to every word in the vocabulary, and whose values are 1 only for those words that are present in the input context and 0 otherwise), and producing a vector as output that, for each word in the vocabulary, provides a probability that it is one of the missing words. In training for this task, the weights between the input layer and middle layer can be extracted and treated as a word vector (often called *embeddings* when produced in this way).[9] The great insight that Mikolov *et al.* (2013) had was that word vectors produced in this way outperformed count models when tested against meaning similarity tasks and did so with fewer dimensions than was typically required for a count model. Furthermore, a striking feature of embeddings produced by Word2vec and similar is that they quickly proved valuable across a range of Natural Language Processing (NLP) tasks. This is partly seen in the wide variety of evaluation tasks for which embeddings provided great leaps of progress on, such as those included in the GLUE benchmark (Wang et al. 2018). But it is also seen more generally in the complete ubiquity of word embeddings throughout NLP today. The state-of-the-art in tasks like named entity recognition, sentiment analysis, machine translation, word sense disambiguation, and coreference resolution all now employ word embeddings in some form or other. As Tenney *et al.* (2018) state, word embeddings are a "staple tool for NLP".

It may be questioned at this stage whether this neural network approach can really be thought of as part of the distributional semantic approach. Whereas the count-based approach explicitly represents the distribution of each word in terms of co-occurrence with other words (at least prior to any dimensionality reduction), the same is not true here. However, it is important to note that because the network is being trained on raw textual data (text that is completely unmarked, unannotated, and comes with no metadata), really the distribution of word types is, in the first instance, all that the model has access to. As Pennington *et al.* (2014, p. 1535) state: "all unsupervised methods for learning word vectors are ultimately based on the occurrence statistics of a corpus". This is further suggested by work from (Levy and Goldberg, 2014), which shows that Word2vec's skip-gram embeddings work by implicitly factorizing a word-context matrix that tracks a statistical association (pointwise mutual information) between words and contexts.

[9] In taking this approach, Mikolov et al. (2013) were effectively combining the distributional semantic method with existing approaches to capturing semantic representations within a neural network, a project that stretches at least as far back as (Hinton, 1986). See: (Bengio, 2008) for further discussion.

The next important technical development was the move towards *transformer architectures* (Vaswani *et al.*, 2017). To understand the need for such architectures, we should first note that the types of embedding produced by Word2vec are *static* in the sense that each word is represented by the same vector no matter the wider text that it appears in. A sentence or phrase is just represented as a sequence of embeddings that otherwise bear no relationship to one another. Furthermore, there is no disambiguation of words: "bank" will be represented by the same vector regardless of whether it is used to talk about rivers or financial institutions. There is a clear need, then, to process text in a way that is sensitive to the words that surround it. A standard way to process a datum in a manner that is sensitive to its surround is to use something like a recurrent neural network. Such networks use hidden states to process information sequentially, so that each point in a sequence is processed in a way that is sensitive to the hidden state generated by the previous point in the sequence. However, these architectures are not well-suited to capturing long-range dependencies in a sequence,[10] and furthermore the sequential nature of the processing (i.e. each word is processed one-at-a-time rather than processing all words at once) means that the training of such models becomes much more computationally expensive.

[10] The recursive nature of language means that, even with a sentence, dependencies between words (e.g. co-reference) could be of any length. For example, in the following sentence, "Mischa" and "he" are co-referential, no matter how many fruits we add to the list: "Mischa went to the shop and bought an apple, an orange, a banana, some grapes, yet he still forgot the strawberries". As the distance between the two expressions increases, recurrent neural networks will struggle to detect this dependency.

Transformer architectures avoid both problems. They are able to capture long-range dependencies in a sequence while also avoiding sequential processing. They do this using self-attention heads, which work in the following way.[11] For each word as represented by a static embedding, we want to replace that embedding with an embedding that is sensitive to its textual surround. A way to do that is to generate a new embedding as a weighted sum of all the context word embeddings that surround the target.[12] How much each word contributes to the weighted sum is determined by an attention score. For every target word, attention scores for each context word are generated by calculating the dot product between the target word and context word.[13, 14] Crucially, each word embedding is playing three distinct roles here: as the target word, as a context word, and in contributing to the weighted sum that produces the new embedding. The embedding goes through distinct linear transformations before being used in each of these roles.[15] In being able to adjust these linear transformations through training, the self-attention head can become sensitive to features of a text, e.g. syntactic or semantic relationships between words. Transformer architectures have multiple layers of multiple self-attention heads, meaning that the self-attention procedure described above can be repeated many times, with the various self-attention heads within each layer focusing on different features of the input text. The result of this process is a contextualized word embedding that is highly sensitive to many different features of the textual context that the word appears in.

[11] More recent work has focused on the importance of the feed-forward networks that the embeddings are passed through after each self-attention layer (e.g. Geva et al. 2021). These layers were originally introduced to add the possibility of non-linear processing, as all processing in the self-attention heads are linear.

[12] Here, "target word" refers to the word for which we are producing a new embedding, and "context word" refers to the words that surround the target word.

[13] The dot product is a similarity measure between two vectors. Cosine similarity is derivable from the dot product together with the magnitude of the vectors.

[14] The dot product values are then run through a softmax function so that they sum to 1 and can be used in the weighted sum for the new embedding.

[15] These roles are known as the query, key, and value, respectively.

The crucial linguistic point to note about transformers is that the architecture does not depart from the basic distributional approach insofar as they take the word embeddings that are at the heart of the distributional approach and then modify them on the basis of their contexts, where the context itself is only represented in distributional terms. That said, there are ways in which LLMs can depart from the basic distributional approach, as models like Chat-GPT do. What I have described thus far, particularly in the overview of transformers, is the training procedure often referred to as "pre-training". Pre-training is usually the most significant training step whether measured in terms of computational power, time, or cost. LLMs can then be "fine-tuned" in a further training procedure, where this may introduce new information. For instance, a model designed to answer questions on a particular topic could be trained on a supervised sample of questions and answers. Or in *reinforcement learning through human feedback* (RHLF) human users can interact with the model and then assess the output, where this information can then be used to fine-tune the weights of the model towards outputs that receive positive feedback.

### *2.1. Skeptical arguments*

With this overview of the construction of LLMs in place, we are now in a position to consider certain initial arguments in favour of the skeptical position. One such argument will appeal to the "language modelling" training task that is used to train LLMs. The thought is that language models are trained merely to predict missing words, and continue to do this when they are used to generate new text, and this is sufficient to establish that LLMs are blind to the meanings of the words they process over. In this respect, LLMs are no different to simpler n-gram models, that predict which word is missing purely on the basis of the n number of words preceding.[16]

[16] Gary Marcus' (2022) line that "large language models are little more than autocomplete on steroids" is indicative of this kind of consideration.

However, this argument ignores the key insight behind the vector space approach discussed earlier – that those embeddings seem to serve as good representations of word meanings across a vast range of NLP tasks related to meaning, and that this in turn serves as a partial vindication of the distributional hypothesis. The important point is not merely that LLMs excel in their training task, but that they provide the basis for the state-of-the-art across a whole host of other meaning-related tasks. To the extent that word embeddings continue to be the state-of-the-art across all such tasks, the distributional hypothesis – which goes beyond a claim about the best way to predict plausible text – is made plausible. In this light, merely pointing to the training procedure is not sufficient to establish the skeptical position.

A similar form of argument starts from the fact that LLMs only have unmarked text available to them in training.[17] Only very basic forms of pre-processing are typically done to this data, such as lower-casing, lemmatization, and splitting the text into sentences, words, and word parts. The argument focuses on this point, claiming that raw textual data will only ever provide information about the distribution of words across that text. As distributional properties are not semantic properties, this approach has no access to the semantic properties of those words. Again, the thought is that in recognizing distributional patterns in the training data, LLMs are able to produce text that is statistically plausible but meaningless. It must be meaningless simply because it was not trained on data that contains insights into the semantic properties of words.

However, this is too quick. It ignores a basic point about what the outcome of a statistical analysis of a dataset can be: that often such an analysis is best understood in terms of the search for latent variables. In the case of vector spaces, the employment of vectors within a high-dimensional space is an opportunity to capture the patterns within the original data in terms of the positions in the vector space, where positions in the vector space may themselves capture value assignments for certain latent variables. Those latent variables may be meaning properties.

[17] This kind of consideration plays a role in Bender and Koller (2020, 5185): "we argue that the language modeling task, because it only uses form as training data, cannot in principle lead to learning of meaning".

Here are two illustrations of the point. First, consider the so-called *analogy findings* from Mikolov, Yih, and Zweig (Mikolov, Yih and Zweig, 2013). They found that certain syntactic and semantic features are encoded into the vector space insofar as pairs of expressions that differ from one another only with regard to one of those features will be offset from one another in the same direction in the space. To use the most well-discussed example, "man" is offset from "woman" in roughly the same direction and distance that "king" is offset from "queen". So it seems that something like [+masculine] and [+feminine] has been represented in the word space even though the raw data that the model was trained on obviously contained no explicit information on the relationship between expressions like "queen" and "women" and such semantic features.

Apart from the analogy findings, it is already widely-recognized that in performing the prediction task in pre-training, LLMs become sensitive to linguistic features that are not explicitly represented in the training data. A common way to investigate this possibility is to produce *probing classifiers* i.e. see whether a supervised network can be trained that would take word embeddings as input and provide as output a verdict on whether some linguistic property is present (Conneau *et al.*, 2018; Tenney, Das and Pavlick, 2019; Miaschi and Dell'Orletta, 2020). To the extent that a network can be trained on such a classification task, it is good evidence that the embeddings have encoded for the property in question. For instance, using this probing technique, Tenney, Das, and Pavlick (2019) found that the LLM BERT encodes information about part-of-speech categories, syntactic constituency, semantic role labelling, and co-reference. In transformer architectures like BERT, this probing technique can be extended to investigating the individual self-attention heads. As each self-attention head outputs a vector, the outputs of each can be tested using a probing classifier. We can then explore whether each self-attention head is sensitive to particular linguistic properties (for an overview, see: (Rogers, Kovaleva and Rumshisky, 2021).[18] While this opens up an important project in terms of reducing the opacity of LLMs, the important point here is that there is already ample evidence that LLMs are able to encode a wide array of linguistic features that are not directly observable in the training data. That it can do this is perhaps a surprising finding, but it is at this stage difficult to deny. Together with the analogy findings, this illustrates how it cannot be inferred from the nature of the training data alone that LLMs have no access to the semantic properties of a language.

But there is still a further argument in favour of the skeptical answer to the meaningful usage question. This argument can concede that LLMs can access some semantic properties, as the various probing classifier studies serve to show. However, there are certain semantic properties – *referential* properties at the level of the expression and *truth-conditional* properties at the level of the sentence – that LLMs must have no access to. So while LLMs may be sensitive to certain semantic properties (perhaps that "queen" has [+feminine] associated with it) and that this may even manifest in terms of an entailment relation (e.g. "x is a queen" entails "x is not a king" or similar), an LLM cannot be sensitive to the fact that there are semantic relationships that hold between words and sets of objects, and between sentences and states of affairs. Word-to-world relations are just not going to be accessible for LLMs.

For many, this is essentially a new version of the problem illustrated by Searle's (1980) Chinese room thought experiment, or what Harnad (1990) dubbed the "symbol grounding problem". There are important differences between Harnad and Searle on this matter that bear mention. Searle was primarily interested in whether a computer could *understand* the information it processes, where Harnad was interested in what it takes for a symbol processing system to enter a *meaningful state*. Searle takes the problem to apply to any computational system whereas Harnad takes the problem to apply to systems that operate over symbolic representations (such as linguistic representations). But both are agreed that a computational system that operates over linguistic representations will have no access to the referential relations that those symbols possess, just as the figure in the Chinese room has no access to worldly relations of the Chinese symbols that he operates over. The argument that we are currently considering carries that problem over to LLMs, in what Coelho Mollo and Millière (2023) dub the *vector grounding problem*.[19] Just as a look-up table of input/output relations over Chinese symbols tells you nothing about the relations between those Chinese symbols and the world, a vector space will not contain any information about how words represented as points in that space bear relations to worldly entities.

[18] Another method of investigating self-attention heads used by (Clark *et al.*, 2019) is to take as output of each head the word that received the highest attention score, and then see whether the output consistently bears some syntactic relation to the inputted word.

This is an important problem that has already generated a great deal of discussion within the NLP community (Bender and Koller, 2020; Michael, 2020). As stated earlier, a natural way to respond to this problem would be to take our best theory of what it takes for an entity to be meaningful in a sense that allows for worldly relations and then consider whether LLMs meet the conditions that the theory provides. In the following section, I will consider Coelho Mollo and Millière's recent attempt to apply a mental metasemantics.

## 3. Mental metasemantic theories and large language models

Before focusing on Coelho Mollo and Millière's specific proposal, it is worth considering again the general shape of metasemantic theories. Metasemantic theories are concerned with how meaning properties are determined. As we saw earlier, there are mental metasemantic theories that attempt to capture how cognitive states come to have the meaning that they do e.g. how exactly does a visual experience, memory, or even a sub-personal representation come to be about whatever it is about? The project can be viewed as an attempt to draw a distinction between those cognitive states that have meaning properties and those that lack them. Then there are linguistic metasemantic theories that attempt to capture how it is that words, sentences, and utterances come to have the meanings that they do. Note a disanalogy between the two projects that will be important in what follows. In the linguistic case, we have a distinction to draw between words and sentences on the one hand, and utterances of words and sentences on the other. A linguistic metasemantics will give some story of how each type of object comes to have the meaning that it does. And of course, there will be a great deal of controversy over how exactly this should be done. Cases of polysemy, context-sensitivity, and pragmatic forms of meaning suggest that utterances do not inherit their meaning wholesale from the uttered sentence. But giving an account of what needs to be added to or subtracted from sentence meaning to determine utterance meaning is a core area of debate across a wide range of expressions. While some form of equivalent distinction arguably does hold in the mental realm between concepts on the one hand and applications of concepts on the other, the distinction is not a core feature of mental intentionality insofar as there are forms of mental intentionality that do not rely on it e.g. perceptual experiences.

[19] Harnad (2024), in dialogue with GPT-4, argues that LLMs still suffer from the symbol grounding problem.

As Burge (1992) noted, the attention paid and progress made in each metasemantic project arguably took place at different periods of the 20th century. This is not the place for a history lesson, but it is true enough to say that particular attention was paid to linguistic metasemantics in the work of Russell, Wittgenstein, Kripke, Putnam, Davidson, Dummett, Evans, with many of the seminal texts being written in or before the 1960s.[20] By contrast, mental metasemantic theories really began to be developed in earnest after the so-called cognitive revolution initiated by Chomsky (1959), and they are perhaps best exemplified in the works of Dretske, Fodor, Millikan, Papineau, and others. So after the available linguistic metasemantic positions started to calcify, attention turned to how something like a cognitive state could have intentional properties. And of course, mental metasemantic theories have played a central role when considering the possibility of artificial intelligence, as it seems a plausible enough claim that if a system is to count as exhibiting artificial intelligence in the strongest sense, it must at least be capable of entering into meaningful states.

[20] There will be many exceptions to this generalization, but one particularly worth noting is Brandom's (1994) work.

Perhaps then, because mental metasemantics has enjoyed greater focus in recent times, and because it has a clear connection to questions in artificial intelligence, it is *prima facie* natural to think that if we are to apply a metasemantics to LLMs, we should turn to mental metasemantic theories.[21]

This has arguably been the dominant approach in early philosophical discussion about LLMs (e.g. Butlin, 2021; Grzankowski, 2024). For the purposes of this paper, I will focus on Coelho Mollo and Millière's (2023) proposal, as I take it to be the most sophisticated philosophical treatment of the topic. They identify two widely-accepted conditions for some internal state of a system to be intentional. First, there must be some causal-informational link between the meaningful state and the external entities that the state is about. Second, they endorse a teleosemantic condition whereby the system must possess some world-involving function, given the history of the system (either the evolutionary history or the individual history, or both). According to teleosemantics, it is because the system possesses a world-involving function, selected in the course of the history of the system, that states of the system can be thought of as having correctness or satisfaction conditions – a key feature of intentionality. To use a well-worn example, it is because the frog has a particular evolutionary history that it can be described as a system that serves a function to catch flies using its brain, eyes, and tongue. So when it shoots its tongue at a speck of dirt, the fact that it has that function for capturing flies is what gives us grounds to say that it misrepresented the dirt for a fly.[22]

[21] A further reason for thinking that we should turn to mental metasemantic theories is some prior commitment to the view that the only form of non-derived intentionality is mental intentionality Coelho Mollo and Millière gesture towards this kind of view when they state "A subset of these mental and cognitive states plausibly enables external public representations – such as words and sentences – to be meaningful. The meaning of these external representations derives from the beliefs, intentions, and conceptual representations of language users" (2023, p. 14). In §4, I will discuss the idea that mental intentionality is more fundamental than, or at least necessary for, linguistic intentionality in greater detail.

Coelho Mollo and Millière consider whether LLMs can be said to meet those two conditions. Regarding causal-informational relations between LLMs and things in the world, the question is whether an LLM could stand in an appropriate causal-informational relationship with things in the world in order to be said to represent them. For instance, if Chat-GPT uses the word "Birmingham" in a sentence about the city, does Chat-GPT stand in an appropriate causal-informational relationship with the city in order to do so? Their answer is yes. Although LLM only ever causally interact with the world in a way mediated by previous language use (i.e. the text that it is trained on), this is really no different to the kinds of causal-informational links that we often have with things that we can talk about. For instance, the only causal-informational links I have to Aristotle are mediated through historical text and testimony. But this does not prevent me from successfully talking and thinking about him. The difference between us and LLMs is that their causal relations to things in the world are always mediated in this way, whereas we are able to enter more direct causal relations with objects, most obviously through perception.

The question then becomes whether LLMs can be said to possess the world-involving functions that the teleosemanticist claims are necessary for any form of satisfaction condition. As Coelho Mollo and Millière note, there is some *prima facie* promise to the idea that they do indeed possess a function of this kind insofar as they have been through a learning procedure in their training. In doing so, the way in which they process and respond to inputs has been selected in their history. But the difficulty in applying this condition to LLMs is in whether there is any plausibility to the idea that the resultant function could be appropriately *world-involving*. That is, for the teleosemanticist to claim that LLMs are able to talk about the world when they output what they do, they will need to claim that LLMs have functions directed at particular entities in the world and that representations of the world help fulfil those functions.

[22] To be clear, the claim is not that any system that serves a function will thereby possess intentional states. According to Millikan – one of the chief proponents of this kind of teleosemantic condition – intentionality comes into the picture when a particular way of serving a function is adopted. The way of serving the function must be suitably world-involving , including some mapping between the internal state and some external state of affairs.

It is arguably difficult to find such a function in the pre-training procedure for LLMs. As we have already seen, the task that LLMs are set in pre-training is usually just one of missing word prediction. Indeed, this led Butlin (2021), while also employing a teleosemantic picture, to be skeptical that LLM outputs could be meaningful.[23] The thought is that the prediction task is merely *intralinguistic* i.e. it is a task focused on probabilistic relations between words, and so is not focused on features of the world.

Coelho Mollo and Millière seem sympathetic to this idea that language models that have only gone through pre-training do not seem appropriately hooked up to the state of the world to be thought of as possessing world-involving functions. However, they argue that there are three possible (and actualized) features of LLMs that they argue could provide such a basis. First, as noted earlier, LLMs are often fine-tuned on some more specific task, where new kinds of information can be introduced. For instance in RHLF users of a chatbot may be asked to give feedback on whether what the chatbot produced was true (or whatever desired feature it is that we want the chatbot to exhibit). This information can be used in fine-tuning, where the model is tasked with tuning its parameters so that it is more likely to produce outputs labelled as true. In this way, Coelho Mollo and Millière suggest, LLMs are being asked to align their outputs with the state of the world, and so could be viewed as gaining a world-involving function through this training procedure.

---

[23] Butlin has since modified his position on the matter and presented his modified position publicly. For the purposes of this paper, all reference to Butlin picks out the pre-2021 time-slice that endorsed the view in the 2021 paper.

They also suggest that LLMs may be able to gain world-involving functions on the fly when they are given “in-context” (i.e. one-or-more shot) learning prompts. This is where a model is given a prompt to complete some task together with an example of that task being completed. Coelho Mollo and Millière suggest that this kind of prompt provides the model with a standard of correctness which goes beyond the proximate task of merely next-token prediction. Finally, they also suggest that there could be quite particular domains of discourse for which the model has been able to infer the standards of correctness through its training. For instance, they suggest that LLMs may be able to learn the rules of a game provided that there are enough recordings of gameplay in the training set. GPT-4, for example, can play tic-tac-toe without any explicit instruction of the rules.

My aim in this section is ultimately to endorse the point made explicitly by Butlin and that is acknowledged by Coelho Mollo and Millière: that a LLM that has only undergone pre-training on next-token prediction has not plausibly undergone the appropriate learning procedure to be thought of as possessing the kind of world-involving function that is typically thought to be required for mental intentionality. After all, if a LLM was trained on a highly sophisticated artificial language that was completely uninterpreted (i.e. none of the expressions possess any worldly relations, referential or otherwise), the model would plausibly do an impressive job of next token prediction and would be able to produce new text that patterned well with the original training data. But I will briefly comment on the plausibility that certain fine-tuning procedures may provide the appropriate basis to claim that LLMs could possess a world-involving function. The problem with this idea is that it would be odd to claim that a relatively marginal procedure that is designed to optimize the performance of a LLM on specific tasks ends up being the difference between intentionality and its lack. The reason, simply put, is that the fine-tuning procedures – whether RLHF, supervised training, or something else – are not the reason why we have such impressive LLMs today. Anyone who has used one of the larger pre-trained LLMs (e.g. GPT-3) as a chatbot prior to fine-tuning can attest to this fact, but this is also made clear in the leaps of progress that have come with the introduction of new pre-trained models even before they are fine-tuned. For instance, the introduction of GPT-3 represented an improvement in the state-of-the-art across a range of NLP tasks even when compared against earlier fine-tuned models (Brown *et al.*, 2020).[24] Note that the claim here is not that LLMs will only be achieved in systems that reach a certain level of performance in NLP tasks. The thought is instead that given that it is the impressive performance of this technology that is leading us to consider the question of whether they can be in meaningful states, it would be odd to claim that such systems can be in meaningful states if they have gone through a relatively marginal training procedure that we have had available to us for quite some time. As Lederman and Mahowald (2024, 8) state "while RLHF influences model capabilities, even models without RLHF produce apparently meaningful text." But when we turn to consider what is responsible for the impressive performance – the development of a system through pre-training that employs word vectors ultimately derived from distributional properties – I agree with Butlin and Coelho Mollo and Millière that such a system does not plausibly satisfy the conditions for a teleosemantic mental metasemantics.

## 4. Linguistic metasemantics and large language models

In section 3, I argued that the attempt to apply mental metasemantic theories to LLMs leads to a negative verdict, for the distributional system that is at the core of LLMs does not meet plausible criteria associated with such theories. Given the difficulty of applying mental metasemantic theories to LLMs, in this section I want to instead consider whether the prospects for applying a linguistic metasemantics fare any better. I will argue that they do. However, it is important that we do not view the project as just trying various forms of metasemantics until we find one that fits. There is good reason for turning to a linguistic metasemantics over a mental metasemantics.

[24] I expect the same is true of the introduction of GPT-4. Unfortunately (for the point made here), the technical paper on the abilities of GPT-4 focused on a model that had already undergone RLHF (OpenAI, 2023)

After all, LLMs obviously only process linguistic information and we then interact with LLM-powered chatbots linguistically, and it is their sophistication in that activity that leads us to consider whether they produce meaningful outputs. In doing so, we are considering whether we should group them with the ordinary language user or with the parrot. But it is linguistic metasemantic theories that tell us what it is to be an ordinary language user – what it is to produce an utterance that is in a particular language. It is, then, linguistic metasemantics to which we should turn.

It is important at this stage to flag again that a linguistic metasemantics has two job descriptions. On the one hand, it aims to capture how words and sentences have the meanings that they do. On the other, it aims to capture how linguistic events i.e. utterances come to have the meaning that they do. As alluded to earlier, there are a whole range of phenomena (ambiguity, context-sensitivity, implicatures) that suggest that the meaning of the utterance is not just the meaning of the uttered sentence. In thinking about LLMs, we are considering not merely whether they are outputting words that themselves have meaning, but whether what they output is a meaningful usage as when an ordinary language user utters a sentence.

It might be thought that once the broader move of this paper is made – the move from considering mental metasemantics to considering linguistic metasemantics – then the possibility of a positive answer to the meaningful usage question becomes eminently more plausible, as a linguistic metasemantics will not require that we attribute intentional mental states to such systems. There is something to this thought, but we need to be careful, for it is a common enough idea that linguistic intentionality relies in some way on mental intentionality.

One way in which this claim might be understood is via some Gricean theory of meaning where sentence meaning reduces to speaker meaning, and speaker meaning reduces to speaker intentions. But it is important to note, as Borg (2008, pp. 251–254) has argued, that we could allow that Grice is in one sense right about that reductive claim without thinking that all speakers thereby have to employ speaker intentions on every occasion of use. It may well be that speaker intentions are "preconditional" to linguistic conventions insofar as the story of how linguistic conventions come about appeals to speaker intentions. A speaker initially intends to use a sign with a particular meaning, and this practice is picked up by the rest of the community, and so a linguistic convention is set. But notice that this says nothing about what it takes for the convention to be followed once it is in place. And in considering LLMs, we are really considering whether they successfully follow linguistic conventions as an ordinary speaker would.

But still, even when considering particular attempts to employ a linguistic convention rather than the formation of those conventions, there is still the possibility that some mentalistic condition holds on the successful production of meaningful linguistic tokens. To see this, consider Kripke's (1980) highly influential causal-historical theory of names. One of the striking features of Kripke's account is that it says so little on what is required to successfully use a name.[25] The core of the view is that to use a name successfully, the speaker must stand in an appropriate causal chain of usage for that name, which will stretch all the way back to the original naming event. But importantly for our purposes, Kripke also required that speakers, in using a name, must *intend* to refer to the same object that previous users of the name referred to – sometimes called a "reference-preserving intention" (Kripke, 1980, p. 96).[26] If successful use of words requires this sort of intention, then it seems we are back to considering whether LLMs can enter into meaningful mental states. Worse still, even if we could allow that intentional states are possible for LLMs, we would also need reason to think that they are producing reference-preserving intentions in particular.

[25] It is partly for this reason that Kripke was hesitant to describe his view as a theory (Kripke, 1980, p. 93).

[26] The requirement is initially added to deal with cases where names are re-used to name a new object. If I name my dog after Castro, then the worry would be that each subsequent use of the name to pick out the dog actually refers to the Cuban leader. Kripke effects a break in these two causal chains with the requirement of a referential intention. When I first use "Castro" to refer to my dog, I do not intend to use it in the way it has previously been used.

The problem generalizes beyond Kripke's account of names. For there are many types of expression where a production of the sign does not seem to be sufficient to determine a particular expressed meaning. Common names, ambiguous expressions, context-sensitive expressions, syntactically-ambiguous sentences all pose the general problem of a sign being associated with more than one meaning.[27] For each sign, one plausible solution will be to claim that the token must be accompanied by a communicative intention that will settle which meaning was expressed by the production of the sign. In this way, our linguistic metasemantics may just lead us back to a mental metasemantics.

Against this thought, in this section I will show that there are plausible linguistic metasemantic theories that would plausibly attribute meaning to LLM outputs without requiring some communicative intention. The theories I will focus on are Evans' (1982) account of naming practices and Millikan's (1984, 2004, 2005) teleological metasemantics. It will be important to emphasize the precise aim of this section. One approach to making the case for a positive answer to the meaningful usage question would be to argue that according to a particular metasemantic theory, LLMs are indeed meaningful users. This would be to tie such a case to the plausibility of that metasemantic theory. The problem with doing so of course is that there is not really a consensus on various questions regarding the proper shape of a metasemantics. Instead, I hope to show not only that there are metasemantic theories that would include LLMs as meaningful users, but also identify what unites the two theories. I will argue that both theories place particular importance on the idea that linguistic intentionality relies on a pre-existing meaningful system, and so both theories recognize that LLMs are able to employ this pre-existing system in the same way that ordinary speakers do. And indeed, it is one of the lessons of semantic externalism that the way in which uttered words belong to a pre-existing system may prove to be more important than the psychological features of the speaker. It is for this reason that Mandelkern and Linzen (2024) have argued that LLMs could be meaningful users, and in this section I will develop that positive case further.

[27] I am using "sign" here to pick out the object identified by its phonological or orthographic properties. I'm not using "word" or "expression" because for some of the phenomena listed here, how we individuate words across various interpretations is controversial. "Bank" might plausibly be thought of as two words rather than one, whereas a polysemous expression like "book" is best thought of as one word with multiple meanings. I use "sign" here to avoid questions about word individuation.

### 4.1. *Evans' account of naming practices*

Gareth Evans (1973, 1982) was critical of Kripke's particular appeal to reference-preserving intentions, and used his famous Madagascar case to illustrate this. This led him to consider more sophisticated ways in which speakers may employ a name without appealing either to Kripke's reference-preserving intention or the kind of descriptivism that Kripke had refuted. In his later work, he (1982) argued that for each name there is a *naming practice* – a body of information that is publicly associated with the named object. While the information is about the named object, it may still be the case (as we are fallible creatures) that some of the information that forms part of the practice is not true of the object.[28,29] What then makes it the case that a naming practice picks out a given individual, if it is not that the individual uniquely satisfies the associated information? To answer this question, Evans distinguishes between *producers* and *consumers* of the naming practice. Producers are those who are able to add information to the naming practice because they have other causal-informational channels open to the object in question – they can directly interact with the object and have demonstrative and recognitional capabilities regarding the object. The community's naming practice then gets updated by the interactions that producers have with the named object. But crucially, not all users of the name interact with the named object in this way. Consumers of the name are able to use the name by storing the information that forms part of the naming practice. We are all consumers of names of the dead. I became able to refer to Aristotle in using the name by engaging appropriately with the naming practice that is active within my linguistic community. Note that although consumers are dependent upon producers for there to be a naming practice at all, they are still users of the name in the fullest sense.

[28] This idea - of a body of information that is not about an individual in virtue being true (only) of that individual – has since become associated with the mental files approach argued for by Recanati (2012) and many others. However, I have focused on Evans' precursory position as he is more focused on the way that a file (or a naming practice) might belong to a linguistic community rather than to an individual. Furthermore, we will see that his distinction between consumers and producers provides a useful way of describing LLMs in relation to ordinary speakers.

[29] Cappelen and Dever (2021, ch. 5) also consider the application of Evans' version of a mental files position to an AI system. However, their focus is on the basic mental files claim, suggested in Evans (1973), that information could be about an object for causal reasons rather than descriptive fit. By contrast, I am focused on the precise account of naming practices one finds in Evans (1982).

In introducing the notion of a naming practice, Evans was not seeking to rid speaker intentions from his view completely. Instead, he only took exception specifically to Kripke's appeal to reference-preserving intentions in order to determine how a name has been used. He argued that it is a necessary condition on the use of a name that the speaker manifests the intention to engage in a particular naming practice (Evans, 1982, p. 387, see also pp. 68, 318-320). But in introducing the notion of a naming practice as a substitute for Kripke's reference-preserving intentions, Evans thereby developed a position that may not have to appeal to speaker intentions. Here I will consider a version of the Evansian picture that appeals to naming practices without subsequent appeal to speaker intentions. The primary reason why Evans thought it necessary to appeal to speaker intentions was to capture disambiguation i.e. interpretation of a sign associated with more than one meaning. This is an issue that I will return to in the concluding section, but it is important to note that there are proposals friendly to the Evansian picture as considered here that would capture disambiguation. For example, Michaelson (2023) has argued for *futurism*, according to which the reference of an uttered word can be determined by subsequent events after the utterance e.g. the way that the utterance patterns with subsequent discourse. This would bolster the Evansian picture with further resources to capture disambiguation without appeal to speaker intention.[30]

[30] Thanks to an anonymous reviewer on pushing me to clarify this point.

LLMs can plausibly be thought of as consumers of naming practices. They are able to use names and other expressions because through their training they have internalized the information associated with each expression. Why would we think they have succeeded in doing so? Simply because LLMs do have associated with each expression a body of information, as shown both in the sophisticated manner of usage and also the explicit answers that they can provide in response to questions. The answers that they provide, coupled with the fact that they are able to produce those answers in a flexible way, rephrasing and expanding upon if necessary, frankly makes it hard to deny that the kind of information that forms part of a naming practice has been stored. Particularly illustrative here is the natural language processing task known as *named entity recognition* (NER). This is the task of identifying and labelling objects of certain types referred to within a text. For instance, in the sentence "Barbara came to the UK to work in the NHS", the task would be to identify that "Barbara" picks out a person, "UK" picks out a location, and "NHS" picks out an organization. While much of the focus within NLP has been on fine-tuning models specifically to complete this task, perhaps by providing a supervised training set or similar, some have studied the ability of pre-trained language models to complete NER tasks, with some impressive early results (Davody *et al.*, 2022; Epure and Hennequin, 2022). Specifically, both studies investigated the extent to which language models would be able to complete a prompt that would reveal a verdict for a NER task, with promising results for models such as BERT, ROBERTA, and GPT-2. The information probed for in such tasks is precisely the kind of information that Evans would argue is included as part of a naming practice.

Notice that we get a pleasingly attenuated result if we adopt the picture of LLMs as consumers. LLMs are quite unlike human speakers insofar as they are only ever consumers of linguistic practices. This nicely captures the idea that there is an important linguistic activity, where the relation between word and world is maintained, that LLMs do not take part in. In some sense, the spirit, if not the letter, of the skeptical answer to the meaningful usage question is vindicated. LLMs are meaningful users of a language but are parasitic on human language users, as all consumers are. In this way, Evans' picture arguably opens up a nice middle ground for LLMs to reside in.

Evans acknowledged that the distinction may be "inexact in some respects" (Evans, 1982, p. 377). One such respect is in the idea that consumers can effect no change on naming practices. This vision is idealised in that it is not clear that there couldn't be situations where consumers do have an effect on the naming practice, provided that they exert enough influence. For instance, imagine that an influential ancient historian succeeds in circulating a falsehood about Aristotle. In that way, it would not be impossible either for LLMs to effect a change in a naming practice provided that they could exert the requisite influence. But the important insight behind the producer/consumer distinction is that naming practices facilitate highly-mediated linguistic representation of an object, even though the very same naming practices are made possible and maintained by more direct interactions with that object. Those more direct interactions are an important part of linguistic practice that LLMs, as consumers, do not have access to.

It may be objected that LLMs could not be engaging with naming practices, because if they have stored information about anything, it is best understood as probability distributions on expressions appearing in particular linguistic contexts. But this objection really returns us to a point made in section 2. Put in somewhat different terms now, the question is really, *conditional upon what* are those probabilities generated? If an LLM is able to access facts about the meanings of particular expressions insofar as the meanings of expressions impact upon how they are distributed across a corpus, then the probability distributions for each expression will reflect that. Again, what is remarkable about the neural network methodology in general, and is particularly salient in LLMs, is the amount and detail of information that can be stored in this way. So just because LLMs are trained on textual data, exploit patterns of distribution, and then are tasked with predicting new text given what has come previously, it does not follow that LLMs could not have exploited information of other kinds that interact with the distributional properties in highly complex ways. The Evansian suggestion here is that it is precisely in this way that LLMs engage with naming practices for particular expressions.

### *4.2. Millikan's teleosemantic account of language*

The Evansian picture discussed in the previous sub-section still places a mentalistic condition on users of a language – in using an expression, a speaker has to process over a naming practice that they have previously stored. I have suggested that LLMs do plausibly meet this condition, thus opening up the possibility that they are meaningful users of a language. In this section, I will turn to an alternative metasemantics that moves even further away from any mentalistic restrictions on language users. Over the past four decades, Millikan (1984, 2004, 2005) has developed a view of language and linguistic meaning that she describes as anti-Gricean insofar as the meaningfulness of a linguistic token is not to be explained by the presence of some mental state. Instead, the meaningfulness of a given token is to be explained in terms of its teleology. Rather than seeking to reduce linguistic intentionality to mental intentionality, she argues that both forms of intentionality can be explained in teleological terms (Millikan, 2005, ch. 5).

According to Millikan, linguistic items are best understood as social conventions. Each item, whether a word, a grammatical mood, or syntactic type, or otherwise can be viewed as having a lineage insofar as the fact that it has been reproduced is partly explained by the fact that it serves some function. Only partly, for there is some arbitrariness to the fact that the linguistic item serves the function that it does; this is the conventional aspect of language. Millikan describes the type of function that linguistic devices serve as "stabilizing functions." Stabilizing functions reside between two systems – in this case, between speaker and hearer – and have the effect of encouraging each system to keep using that device in order to elicit the same response. Simply put, it is good for speaker and hearer that there is the convention of a hearer responding in a particular way to particular linguistic act, and this serves to explain why the convention continues to exist.

The stabilizing functions that linguistic items serve may themselves be cashed out in terms of mental states. For instance, Millikan (2005, pp. 62, 44, 157) repeatedly uses the example of grammatical mood and claims that the declarative mood has the function of instilling beliefs into the hearer. But that a linguistic device has the function of producing true beliefs in a hearer does not mean that on each occasion where the linguistic device occurs, belief production must occur. A linguistic device on a particular occasion can possess a function that it fails to perform; it is the reproductive history of the device that determines the function it possesses, not what it does on any particular occasion.

This much tells us how linguistic items serve functions but we do not yet have an account of linguistic intentionality. To repeat the point made in fn. 23, linguistic intentionality does not occur wherever a linguistic device possesses a function, or even a particular type of function. Instead, intentionality is a particular *way* of serving a function. Millikan's idea is that a way of serving a function is intentional when it exploits a mapping relation between the device serving the function and some external state of affairs (whether real or otherwise). In the much-discussed example, the bee dance performs the function of stimulating a particular kind of bee response in the observing bee, and it does this because there is a mapping relation between the properties of the dance and some external state of affairs (i.e. the presence of nectar). This gives rise to satisfaction conditions – a key property of the intentional. Turning to a linguistic example, a declarative sentence is a linguistic device that serves a particular function – to instil a certain belief in the hearer. It does this in a particular way, by exploiting a mapping between the sentence and the possible state of affairs described, and it is this mapping that gives rise to the truth conditions of the sentence.[31]

[31] The focus here is largely on declarative sentences, and of course it will be crucial that the account can generalize to non-declarative sentences. Millikan has already developed her position in that direction. For instance, in (1984, ch. 3) she argues that the function of the imperative mood is to produce compliance in the hearer.

This much provides an overview of Millikan's account of intentionality, but the crucial point for our purposes is that her view allows for a way in which linguistic intentionality can be understood in terms of its communicative potential without needing to appeal to the production of particular mental states by speaker or hearer. In appealing to the reproductive history of linguistic items, Millikan argues that there is no requirement that the production of a meaningful linguistic token must be accompanied by some mental state. It is the lineage of the token, and not its mental luggage, that accounts for its meaning. Turning to LLM outputs, we need only be able to understand the tokens produced by LLMs as being part of the same teleological lineage in order for LLM outputs to be thought of as meaningful. And this is where the core distributional model described in section 2 seems well-suited. For what LLMs are trained on is essentially a history of linguistic devices being used for particular purposes. In becoming sensitive to where and when linguistic expressions have previously been used, LLMs are designed to track the functions of the expressions and how that will account for future reproduction. On this picture, what separates meaningful from meaningless tokens of a word is that the stabilizing function serves as part of the explanation as to why it has been reproduced on that occasion. In the case of the ant, the stabilizing function plays no explanatory role, for it is only coincidence that the ant has traced the word. But what distinguishes LLMs from, for example, a simple bi-gram model that produces the most likely word given the word that has come immediately prior? The thought here is that the LLM through its pre-training has been able to analyse the factors that lead to the word being reproduced as these factors serve as latent variables in the statistical analysis of the word's distribution. This subsequently informs the LLM decision to output that word on a particular occasion. In this respect, the stabilizing function plays an explanatory role when used by the LLM but not the bi-gram model. Rather, the bi-gram model is only able to perform a much more shallow distributional analysis that will likely not capture facts about why a particular word has been reproduced.

This picture certainly merits further investigation. In particular, it ought to be noted that there seems an important difference between the explanatory role of the stabilizing function in the LLM case to the human case – where in the LLM case the explanatory role is tied to the massive analysis of word usage history as part of pre-training. But perhaps the right thing to say on this score is simply that this highlights that while LLMs meet the conditions for meaningful usage, they achieve that in a quite particular, non-human way.

## 5. Conclusion: context, content, and intentions

In the previous section I have shown that there are plausible linguistic metasemantic theories that will capture the idea that LLM outputs are meaningful, and that whether this is the case may not depend on the presence of communicative intentions. That LLMs look to stand a better chance of meeting the conditions of a linguistic metasemantics partly reflects a difference across the two types of metasemantics in their explananda. Instances of linguistic intentionality are reliant on a pre-existing linguistic system in a way that instances of mental intentionality need not be. Indeed, this is a feature of language that is made much of by both metasemantic theories considered: in Evans' notion that the meaning of name partly consists in the naming practice that circulates around the community, and in Millikan's notion that the meaning of a linguistic item is understood in terms of its history of reproduction. As such, there is a sense in which generating an instance of linguistic intentionality is less demanding. At the same time, we have also seen that there is some plausibility to the idea that communicative intentions might play some necessary role on the production of a meaningful token.

Ultimately, a great deal of further work will need to be done to defend a positive answer to the meaningful usage question. One area where much of the debate around communicative intentions has arisen in more recent years concerns the meanings of context-sensitive expressions, such as indexicals or demonstratives. What exactly determines the reference for a particular token of "that", for instance? Many have appealed to some form of speaker intention, while others have resisted the move (Borg, 2004, 2012; Gauker, 2008; Lewis, 2020, 2022; Stojnić, 2021).[32] But this is an area where a great deal of further work is required in thinking about LLM outputs, for even if the intention-based picture is rejected, the alternatives usually defended often appeal to contextual elements that LLMs do not have much (if any) access to e.g. demonstrations, prosodic contours, perceptual salience, etc.

Regarding the semantic value of demonstratives, Michaelson and Nowak (2022) have recently defended a form of pluralism whereby a given demonstrative can possess more than one semantic value on a particular occasion of use.[33] They have in mind problem cases where what the speaker intended to refer to and what the speaker gestured towards diverge (e.g. Kaplan's (1978) Carnap-Agnew case). They argue that rather than try to adjudicate between those positions that appeal to speaker intentions and those that don't, it is preferable to allow that each position is correct about one of the many semantic values that the token possesses. In doing so, they argue that we should differentiate between "meaning *qua* first-personal mental report and meaning *qua* publicly-accessible object" (2022, p. 153). I will conclude by suggesting that an added attraction of this kind of pluralism is that it has the flexibility to allow for meaningful linguistic entities in the new kinds of discourses we find generated by LLMs, while allowing that speaker intentions do have some important role to play in meaning determination. Perhaps communicative intentions are one route to meaning, but LLMs do not take that route.

[32] Of course, Millikan can also be added to this list, although she hasn't written as extensively on the issue of context-sensitive items. In (Millikan, 2005, ch. 10) she provides an explanation of various context-sensitivity phenomena via a perceptual model of utterance interpretation.

[33] In a similarly ameliorative tone, Lewis (2020, pp. 1528–9) has suggested that while speaker intentions may not be necessary for determining demonstrative reference, they may nevertheless play a secondary determinative role insofar as when they are present, they may play a role.